\documentclass[a4paper]{article}

\usepackage{INTERSPEECH2019}
\usepackage{multirow}
\usepackage{xcolor,colortbl}
\usepackage{hyperref}
\definecolor{LightCyan}{rgb}{0.88,1,1}
\definecolor{Gray}{gray}{0.85}

\title{ Speech Replay Detection with \textit{x}-Vector Attack Embeddings and Spectral~Features }
\name{Jennifer Williams and Joanna Rownicka}
\address{The Centre for Speech Technology Research, University of Edinburgh, United Kingdom}
\email {j.williams@ed.ac.uk and j.m.rownicka@sms.ed.ac.uk}

\begin{document}

\maketitle

\begin{abstract}
    We present our system submission to the ASVspoof 2019 Challenge Physical Access (PA) task. The objective for this challenge was to develop a countermeasure that identifies speech audio as either bona fide or intercepted and replayed. The target prediction was a value indicating that a speech segment was bona fide (positive values) or ``spoofed'' (negative values). Our system used convolutional neural networks (CNNs) and a representation of the speech audio that combined \textit{x}-vector attack embeddings with signal processing features. The \textit{x}-vector attack embeddings were created from mel-frequency cepstral coefficients (MFCCs) using a time-delay neural network (TDNN). These embeddings jointly modeled 27 different environments and 9 types of attacks from the labeled data. We also used sub-band spectral centroid magnitude coefficients (SCMCs) as features. We included an additive Gaussian noise layer during training as a way to augment the data to make our system more robust to previously unseen attack examples. We report system performance using the tandem detection cost function (tDCF) and equal error rate (EER). Our approach performed better that both of the challenge baselines. Our technique suggests that our \textit{x}-vector attack embeddings can help regularize the CNN predictions even when environments or attacks are more challenging.
\end{abstract}

\section{Introduction}
Detecting replayed speech is a notoriously difficult task in speech processing. In particular, an adversary might obtain a snippet of recorded speech wherein the human target has used their voice authentically, such as when speaking a passphrase. When an automatic speaker verification (ASV) system is presented with fake speech intending to impersonate a live talker then this is commonly referred to as either ``spoofing'', a \textit{replay attack} or a \textit{presentation attack}. Building upon similar challenges from earlier years~\cite{todisco2017constant,font2017experimental,lavrentyeva2017audio}, detecting replay attacks was the basis for the Physical Access (PA) task of the ASVspoof 2019 Challenge~\cite{asv2019}. The aim of our work was to detect whether or not a bona fide live speech recording had been intercepted and subsequently replayed back as non-live speech to an ASV system. There are numerous variables to be modeled including elements of how the attack was conducted as well as the type of ASV system being attacked. Our approach captured these conditions using a convolutional neural network (CNN) architecture, similar to the top system from the earlier ASVspoof 2017 replay detection challenge~\cite{lavrentyeva2017audio}. We explored several types of signal features such as those from~\cite{font2017experimental} and our final submission was based on our own specially-trained \textit{x}-vector~\cite{snyder2018x} embeddings combined with signal features.

While there are many different kinds of speech spoofing attacks, including voice conversion and text-to-speech~\cite{janicki2016assessment,shiota2015voice,shiota2016voice,blue2018hello}, we have focused on replay attacks. Some work suggests that speech replay clues are found in the time and frequency domains \cite{lai2019attentive}. Other work has explored energy-based features~\cite{Kamble2018,kamble2019analysis}, attention-based adaptive filters \cite{liu2019replay}, and convolutional neural networks (CNNs) \cite{chettri2018study,himawan2019voice}. It is particularly challenging to model different types of acoustic environments, playback devices, and recording devices \cite{chettri2018analysing}. Recent work has also shown that high-frequency sub-bands in the acoustic signal contain evidence of replay. For example, Inverted Mel Frequency Cepstral Coefficients (IMFCCs) consistently discriminate speech replay across several frequency sub-bands~\cite{witkowski2017audio}. IMFCCs come from inverting filters in the frequency domain to capture more detail from higher frequencies. Other recent work has shown that Sub-band Spectral Centroid Magnitude Coefficients (SCMCs) are the best and most consistent signal feature~\cite{font2017experimental} across experiments, while the Constant-Q Cepstral Coefficient (CQCC) features are also promising~\cite{nagarsheth2017replay}.

This paper makes three main contributions. First, we introduce novel \textit{x}-vector attack embeddings capturing the recording conditions of an utterance (attack and environment variables). Second, we analyze how well the \textit{x}-vector embeddings model factors of variation from different recording conditions. Finally, we demonstrate that the combination of signal features and \textit{x}-vector embeddings out-performs all baselines for both metrics on the ASVspoof 2019 development and evaluation datasets.

\section{Feature Development}
\subsection{Speech Signal Features}
\label{sec:sig_feats}
Following from the features analysis for replay attack detection that was presented in~\cite{font2017experimental} for the ASVspoof 2017 challenge, we extracted the following features: Mel Frequency Cepstral Coefficients (MFCCs), Inverted Mel Frequency Cepstral Coefficients (IMFCCs), Rectangular Filter Cepstral Coefficients (RFCCs), Linear Frequency Cepstral Coefficients (LFCCs), Sub-band Spectral Centroid Magnitude Coefficients (SCMCs), and Constant Q Cepstral Coefficients (CQCCs)~\cite{todisco2017constant}. A description of the features is provided in Table~\ref{tab:sigfeats}. We used static features because preliminary experiments indicated that the dynamic features were not as useful in the spoofing task, especially the second-order features.

\begin{table}[h!]
\centering
\begin{tabular}{c|c|c}
 Features & Coeff. num. ($N$) & $f_{min} - f_{max}$ (Hz)\\\hline
MFCC & 70 & 300-8000\\
IMFCC & 60 & 200-8000\\
RFCC & 30 & 200-8000\\
LFCC & 70 & 100-7800\\
SCMC & 40 & 100-8000\\
CQCC & 50 & 15.62-8000\\
\hline
\end{tabular}
\caption{Description of signal features extracted from the raw speech audio. Note CQCC used a specific $f_{min}$ from \cite{todisco2017constant}}
\label{tab:sigfeats}
\end{table}

We used the IDIAP \textit{Bob.ap} signal processing library to extract this set of signal processing features from speech audio files~\cite{bob2012,bob2017}. In the case of CQCCs, we used the code provided by the challenge organizers and also described in~\cite{todisco2017constant}. For each audio file, the feature extractor output is an $NxM$-\textit{dimensional} matrix with $N$ as the number of coefficients and $M$ as the duration of the file in frames. 

Each audio file in the ASVspoof 2019 dataset was of a different length. We pre-processed the extracted features to handle this length variability and to create same-sized feature vectors as input to our CNN classifier. We used a down-sampling technique in the feature space. This means that for a given coefficient in a given audio file, we down-sampled the number of frames to a constant. The technique preserved the original per-coefficient distributions in a file while also setting the number of frames to a constant. For re-sampling we used the Fast Fourier Transform (FFT) so that the spacing between the original frames, $s=\delta x$, then became: $s=\delta x * M / M'$. By doing this, we set the number of down-sampled frames $M'$ to a constant where $M'=10$. This effectively shortened the audio file duration to 10 frames while preserving the mean and standard deviation of each coefficient in a given file~\cite{scipy}. 

After this, our signal processing features were represented as a $Nx10$-\textit{dim} matrix. We selected $M'=10$ frames based on two motivations: 1) to reduce the overall size and overhead of the dataset for later processing, and 2) to allow our countermeasures to operate on very short audio examples. We then stacked the coefficients on a per-frame basis, in an effort to preserve some temporal nature of the original signal. Finally, each of the signal feature values were re-scaled to be between $-1$ and $+1$ using per-feature max values from the training set, applied later to development and evaluation sets. We re-scaled the values in order to align with our selection of activation function for the regression task described in Section~\ref{sec:system}, which outputs a value between $-1$ (spoofed) and $+1$ (authentic) .

\subsection{\textit{X}-vector Embedding Creation}
\label{sec:xve_feats}
In this work we also used \textit{x}-vectors~\cite{snyder2018x} as auxiliary features for the CNN model described in Sec.~\ref{sec:system}. Our aim was to extract meaningful fixed-size utterance-level vectors representing the factors of variation, namely environment and attack conditions, in the spoofing task. We used our extracted representations to account for these factors in the final spoofing detection task. This effort was to improve the system robustness to unseen conditions by leveraging information about the environment and attack classes from the labels provided for each training example. 

The Kaldi Toolkit~\cite{Povey_ASRU2011} was used to extract \textit{x}-vectors representing a joint environment+attack class (which we refer to as \textit{env+attack}). The input features for the \textit{x}-vector extractor were 40-\textit{dim} MFCCs, with 80 filters in a filter bank. The \textit{x}-vector extractor was a time-delay neural network (TDNN) with the same architecture as in~\cite{snyder2018x}, i.e. seven layers with batch normalisation and ReLU activations. The \textit{x}-vectors were extracted from the sixth layer before the nonlinearity. Differently to the model in~\cite{snyder2018x} though, ours was not trained to classify speakers. The extractor was trained to differentiate between classes jointly representing types of acoustic environments and types of attacks. 

The joint \textit{env+attack} classes were created by combining each category label of variation for the simulated acoustic environments and attack types, i.e. room size, T60 reverberation time, talker-to-ASV distance, attacker-to-talker distance, and replay device quality. From 10 attack type configurations, and 27 acoustic environment configurations (9 attacks plus authentic speech), we created 270 env+attack classes. Training an x-vector extractor to differentiate between \textit{env+attack} classes enabled us to learn fixed-size representations, capturing both the type of attack and the type of acoustic environment. The classification accuracy of our joint \textit{env+attack} \textit{x}-vectors was around 85\% for 270 unique classes on a held-out validation set (10\% of training), hence these representations were meaningful. 

After the \textit{x}-vector extraction, we reduced the dimensionality of our \textit{x}-vectors from 512-\textit{dim} to 10-\textit{dim} using Linear Discriminant Analysis (LDA). The LDA model also used \textit{env+attack} classes for training. We selected 10-\textit{dim} based on the EER from the development set. For 59,400 trials with non-target proportion of 50\%, the EER in the \textit{env+attack} verification task was 23.96\% with the LDA backend.

\subsection{\textit{X}-vector Embedding Analysis}
\label{sec:xve_feats_analysis}

In this section we show an analysis of how the \textit{x}-vector embeddings could differentiate between different types of attacks and different types of environments. There are 27 environment classes and 10 attack classes. It was easier to show the analysis for environments and attacks separately, compared to modeling all 270 classes for the jointly trained \textit{env+attack} embeddings, which are the ones ultimately used in our system. 

Environment classification was an easier task (accuracy 86\% on the validation set) than attack type classification (accuracy 60\% on the validation set), even though the number of classes for classifying attacks was smaller than for classifying environment. We hypothesize that this may be caused partly by data imbalance in the case of attack recognition, compared to the evenly distributed examples for the environment classes. 

\begin{figure}[htbp]
  \centering
  \centerline{\includegraphics[height=7cm]{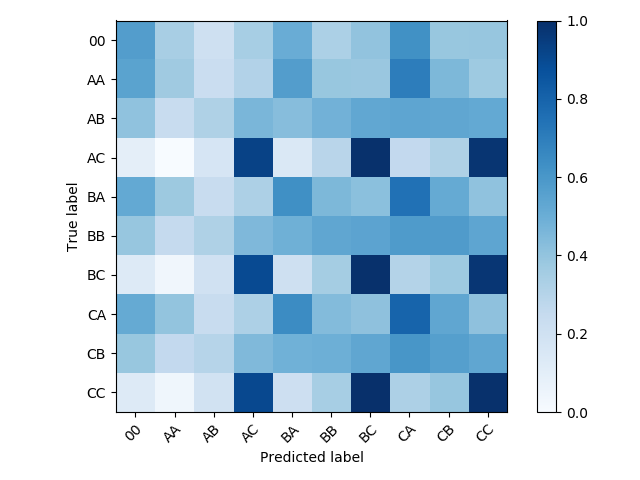}}
  \caption{Confusion matrix for attack class predictions. The scoring is based on the per attack mean development \textit{x}-vectors against per attack mean training \textit{x}-vectors. Class \textit{00} is the bona fide class. For other labels, letters in the first position corresponds to the attacker-to-talker distance (A - lowest, C - highest), letters in the second position corresponds to the replay device quality (A - the best, C - the worst). }
\label{fig:attack}
\end{figure}

To further investigate what the extracted \textit{x}-vectors were capturing, we analyzed the accuracy scores from predicting attack and environment. Figure~\ref{fig:attack} shows a confusion matrix for the scores for mean x-vectors per attack. First of all, it can be observed that the replay device quality is well captured by the \textit{x}-vectors (labels with the same letter in the second position). However, the attack-to-talker distance does not seem to be modeled well with the attack x-vectors (e.g. scores for classes AC, BC, and CC are very close to each other, but different than for any other classes). The most evenly distributed scores can be observed for the medium replay device quality classes (AB, BB, CB). Our \textit{x}-vector embeddings can detect when the replay device is very poor. However, if the replay device quality is near perfect, it is much more difficult to develop the spoofing countermeasures.

\begin{figure}[!ht]
  \centering
  \centerline{\includegraphics[height=7cm]{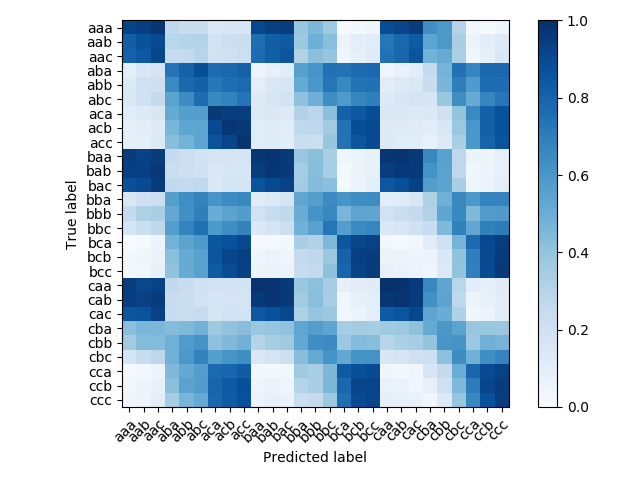}}
  \caption{Confusion matrix for environment class predictions using mean development \textit{x}-vectors against mean training \textit{x}-vectors. Letters in the first position of the label ID correspond to the room size (a - smallest, c - biggest). Letters in the second position correspond to T60 reverberation time (a - shortest, c - longest). Letters in the third position correspond to the talker-to-ASV distance (a - shortest, c - longest). }
\label{fig:env}
\end{figure}

Figure~\ref{fig:env} shows the confusion matrix for how well the mean \textit{x}-vectors discriminate environment classes. Again, the talker-to-ASV distance and the room size do not seem to be very well captured (first and third letter position in labels). However, the reverberation time (especially short and long) discriminates classes well. Both spoofed and bona fide recordings were simulated in a variety of environments. So the idea was to extract an embedding that would help to normalize out the effects of recording in different environmental conditions, to be able to generalize well to unseen conditions at test time.

We hypothesize that even though the \textit{x}-vectors do not differentiate very well between every attack and every environment class, they do differentiate between some of them. Furthermore, \textit{x}-vector embeddings for similar acoustic conditions are close to each other in the \textit{x}-vector space. Therefore, these can be useful representations complimenting our signal processing features.

\subsection{Feature Combination}
In our experiments, we explored several combinations of our features while evaluating on the development set. For the first case, we evaluated signal processing features individually. Next, we evaluated \textit{x}-vector embeddings individually. Finally, we combined signal processing features with \textit{x}-vector embeddings. Before concatenating the LDA \textit{x}-vectors to the signal processing features, the LDA \textit{x}-vectors were scaled with $c=0.1$ constant. We empirically found the scaling to have a good effect on the final EER and tDCF metrics. Scaling \textit{x}-vectors before concatenation is conceptually similar to applying a fixed LDA-like transform in Kaldi (usually used when \textit{i}-vectors are concatenated to the input features for normalisation in ASR), which is scaling down the dimensions that are ``non-informative''. This has the effect of encouraging stochastic gradient descent (SGD) to ignore non-informative values. Scaled and transformed \textit{env+attack} \textit{x}-vectors are denoted as \textit{xEAs} in our paper. They were concatenated to the signal processing features at the input of the CNN model. We hypothesize that it enabled us to normalize out some factors of variation, subsequently enabling the CNN model to learn more robust representations for the final countermeasure task.

Our system submission to the ASVspoof 2019 Challenge was based on two features: SCMC signal features concatenated with the \textit{xEAs} vectors (scaled and transformed as described above). For the submission, our dataset consisted of 54,000 training instances and 29,700 development instances. Our training data was therefore of size $(54000, 410)$. These dimensions were based on 40 SCMC coefficients by 10 frames per coefficient, plus the additional 10-dimension \textit{x}-vectors. The two features were combined via concatenation. In each of the training and development sets, we found 5,400 instances had been labeled as bona fide while the remaining had been labeled as spoof, thus the dataset was very imbalanced for the two targets.

\section{System Architecture}
\label{sec:system}
Our system\footnote{\url{https://github.com/rhoposit/ASVchallenge2019.git}} was implemented with the Keras library~\cite{chollet2015keras} with TensorFlow backend~\cite{abadi2016tensorflow}. We designed our system to perform regression. To do this we converted categorical target labels to numerical values as follows: ``spoof'' became $-1$ and ``bona fide'' became $+1$. Since we used the hyperbolic tangent activation function, the output of our system was therefore a value between $-1$ and $+1$. The challenge evaluation plan called for values with greater negative magnitude to correspond to the ``spoof'' class while values with greater positive value to correspond to the ``bona fide'' class. Thus we intentionally set up the problem as a regression task. Our system output could be interpreted to represent the degree of authenticity of the audio especially considering that the countermeasures output by our system were later evaluated in tandem with an ASV system.

\begin{figure}[ht!]
  \centering
  \centerline{\includegraphics[height=3cm]{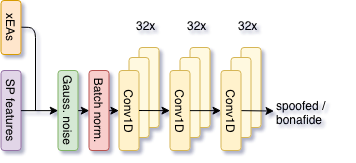}}
  \caption{Overview of our machine learning system architecture. Scaled LDA x-vectors (xEAs) were concatenated to the signal processing features (SCMCs). The DNN was a 3-layer convolutional neural network.}
\label{fig:cnn}
\end{figure}

\subsection{Convolutional Neural Networks (CNN)}
Figure~\ref{fig:cnn} shows an overview of our system architecture. The first layer of our CNN was an additive Gaussian noise layer~\cite{dutta2018convolutional,an1996effects,bishop1995training}. We used this as a form of data augmentation to help the model generalize to unseen conditions. We determined the placement of this layer experimentally and also tried different values for the standard deviation of the noise distribution, finally deciding on $n_{std}=0.001$. The next layer was a batch normalization layer. The CNN consisted of 3 Conv1D layers. The kernel size was set to 3 and we used 32 convolutional filters. Each Conv1D layer included a L2 regularizer~\cite{ng2004feature}. Each Conv1d layer was followed by a max pooling layer with pool size and stride set to 2. Finally, we used a fully connected layer with a single output and the hypoerbolic tangent activation function (\textit{tanh})~\cite{lecun2015deep}. The activation function had the effect of restricting the output between $-1$ and $1$.

\subsection{System Training}
We trained our model using a 10\% subset of the training set as a validation set. During training we swept several different parameters. We explored values for standard deviation in the Gaussian additive noise layer, in the range of: [0.000001, 0.00001, 0.00005, 0.0001, 0.001]. We also explored incremental values for L2 regularization between [0.00001, 0.001]. For training, we measured loss using mean-squared-error (MSE). We used early-stopping~\cite{prechelt1998early}, and monitored validation loss with change $delta=0$ and patience $p=5$ epochs. 
We used the Adam optimizer~\cite{adam} with learning rate $lr=0.001$ and remaining parameters set as default.

\section{Results and Discussion}
Our system was evaluated using two related metrics. The primary metric was the tandem detection cost function (tDCF) computed in conjunction with an ASV system that was kept hidden from participants~\cite{kinnunen2018t}. This allowed the organizers to vary the ASV system to evaluate robustness of systems. The secondary metric was equal-error-rate (EER) based on the quality of the countermeasure alone to predict bona fide versus spoofed. We selected our best system for submission to the challenge based on performance with the tDCF metric.

In Table~\ref{tab:all_results} we report our system performance on the development set using the signal features, the \textit{x}-vector features, and our top features combined. We also report the official results for our submission on the evaluation set. For reference, we provide the evaluation performance for both baselines - which used a Gaussian Mixture Model (GMM) and signal features. They were: LFCC-GMM~\cite{sahidullah2015comparison} and CQCC-GMM~\cite{todisco2017constant}. Our system performed better than the baselines for both metrics on development and evaluation sets (to 3-significant digits).

\begin{table}[!ht]
\centering
\begin{tabular}{lc|c|c|c|c}
 & & \multicolumn{2}{c|}{Development} & \multicolumn{2}{c}{Evaluation} \\
 & & t-DCF & EER (\%) & t-DCF & EER (\%) \\\hline

\parbox[t]{0.5mm}{\multirow{6}{*}{\rotatebox[origin=c]{90}{Signal Features}}}
&\textbf{MFCC}  & \textbf{0.204} & \textbf{8.35} & -& - \\
&\textbf{IMFCC} & \textbf{0.199} & \textbf{7.98} & -&- \\
& RFCC  & 0.210 & 8.58 &-&- \\
&\textbf{SCMC}  & \textbf{0.209} & \textbf{8.47} & -&- \\ 
& LFCC  & 0.229 & 8.90 &-&-\\
& CQCC  & 0.275 & 10.9 &-&- \\  \hline \hline
\parbox[t]{0.5mm}{\multirow{6}{*}{\rotatebox[origin=c]{90}{x-vectors}}}
& \textbf{xA} & \textbf{0.814} & \textbf{31.5} &-& -\\ 
& xE & 0.971 & 41.5 &-&- \\ 
& \textbf{xEA} & \textbf{0.820} & \textbf{31.6} &-&- \\ 
& \textbf{xAs} & \textbf{0.815} & \textbf{31.4} &-&- \\ 
& xEs & 0.970 & 41.7 &-&- \\ 
& \textbf{xEAs} & \textbf{0.820} & \textbf{31.9} &-&- \\ \hline \hline
\rowcolor{LightCyan}  
& \textbf{SCMC+xEAs} & \textbf{0.194} & \textbf{7.74} & \textbf{0.235} & \textbf{9.15}\\ 
\parbox[t]{0.5mm}{\multirow{3}{*}{\rotatebox[origin=c]{90}{Combo}}}
& SCMC+xEAs-N &  0.225& 9.16 & - &- \\ 
& IMFCC+xEs & 0.197 & 7.47 &-& -\\ 
& MFCC+IMFCC & 0.206 & 7.96 &-&- \\ \hline \hline
\rowcolor{Gray} 
& LFCC-GMM& 0.255 & 11.9 & 0.301 & 13.5\\
\rowcolor{Gray} 
& CQCC-GMM & 0.195 & 9.87 &0.245 &11.0 
\end{tabular}
\caption{Our system using different features on the development set, and the evaluation set for our challenge submission. Two baselines from the organizers are included last, for reference.}
\label{tab:all_results}
\end{table}

While the \textit{x}-vectors alone did not distinguish well between spoofed and bona fide speech, we did find an improvement when the \textit{x}-vectors were combined with signal features. Specifically, we found our best feature combination to be SCMC features with the scaled LDA \textit{x}-vectors, \textit{xEAs}, that jointly modeled environment and attack variations. The SCMC feature captures the magnitude of energy a sub-band, which can effectively distinguish two signals even if they share the same average energy. The SCMC feature was also one of the best-performing features from the analysis in~\cite{font2017experimental} for the ASVspoof 2017 challenge, though was based on different data. It has been recognized as a stable feature across experimental conditions. Further experiments on the use of the Gaussian noise layer indicated that using the noise layer (shown by default in Table~\ref{tab:all_results}) always performed better than without it (shown as SCMC+xEAs-N in Table~\ref{tab:all_results}).  

In our earlier analysis, we had found that differences in the replay device quality was captured by the \textit{x}-vectors. When the replay device quality is very good, or perfect, then it is much more difficult to develop spoofing countermeasures. While not reported in this paper, the official detailed performance results from the ASVspoof 2019 challenge organizers also indicate that certain attack types are much more difficult, specifically with the high-quality replay device. That is an important finding for ASV research and our analysis in this paper also supports it. 

A potential limitation of our approach is due to our use of the \textit{tanh} activation function for the CNN regression output. Typically, this activation forces output near the boundaries of -1 and +1, making it more difficult to obtain output scores near the centerline. An interesting analysis of our system might include how the output values are situated within the range of -1 and +1 for various attack types and environmental conditions. 

As future work, we would also like to experiment with different input features for the \textit{x}-vector extraction. In this paper, we used frame-level 40-\textit{dim} MFCC features, without restricting the frequency range. We encourage future work to look more closely at the best frequency range and the best type of frame-level features to capture differences in the acoustic conditions of an utterance with these \textit{x}-vector embeddings. Different normalization techniques could also be investigated.

\section{Acknowledgements}
This work was supported in part by the EPSRC Centre for Doctoral Training in Data Science, funded by the UK Engineering and Physical Sciences Research Council (grant EP/L016427/1) and the University of Edinburgh. It was also supported by a PhD studentship from the DataLab Innovation Centre, Ericsson Media Services, and Quorate Technology. The authors thank: Erfan Loweimi, Ondrej Klejch, and Joachim Fainberg (CSTR), and Michael Camilleri (IANC) for their helpful discussions.

\newpage
\bibliographystyle{IEEEtran}
\bibliography{template}
\end{document}